\documentclass{article} 
\usepackage{iclr2026_conference,times}
\usepackage{amsmath}
\usepackage{hyperref}
\usepackage{url}
\usepackage{algorithm}
\usepackage{algpseudocode} 
\usepackage{graphicx}
\usepackage{booktabs}
\usepackage{todonotes}
\usepackage{amssymb}

\begin{document}

\title{TabSHAP }


\author{Aryan Chaudhary\thanks{Equal contribution.},Prateek Agarwal\footnotemark[1],Tejasvi Alladi\thanks{Advising author.}  \\
Department of Computer Science\\
Birla Institute of Technology and Science\\
India\\
\texttt{\{f20230651,f20230656\}@bits-pilani.ac.in} \\
}
%

\newcommand{\fix}{\marginpar{FIX}}
\newcommand{\new}{\marginpar{NEW}}

\iclrfinalcopy 

\maketitle

\begin{abstract}
Large Language Models (LLMs) fine-tuned on serialized tabular data are emerging as powerful alternatives to traditional tree-based models, particularly for heterogeneous or context-rich datasets. However, their deployment in high-stakes domains is hindered by a lack of faithful interpretability; existing methods often rely on global linear proxies or scalar probability shifts that fail to capture the model's full probabilistic uncertainty. In this work, we introduce TabSHAP, a model-agnostic interpretability framework designed to directly attribute local query decision logic in LLM-based tabular classifiers. By adapting a Shapley-style sampled-coalition estimator with Jensen--Shannon divergence between full-input and masked-input class distributions, TabSHAP quantifies the distributional impact of each feature rather than simple prediction flips. To align with tabular semantics, we mask at the level of serialized \texttt{key:value} fields (atomic in the prompt string), not individual subword tokens. Experimental validation on the Adult Income and Heart Disease benchmarks demonstrates that TabSHAP isolates critical diagnostic features, achieving significantly higher faithfulness than random baselines and XGBoost proxies. We further run a distance-metric ablation on the \emph{same} test instances and TabSHAP settings: attributions are recomputed with KL or L1 replacing JSD in the similarity step (results cached per metric), and we compare deletion faithfulness across all three.
\end{abstract}
\section{Introduction}

The rise of Large Language Models (LLMs) has fundamentally altered the landscape of machine learning, demonstrating remarkable capabilities across diverse domains - from natural language understanding to code generation. Recently, this transformation has extended to structured data analysis, where LLMs fine-tuned on serialized tabular datasets have emerged as viable alternatives to traditional tree-based methods like XGBoost and Random Forests. By converting tabular rows into natural language prompts, approaches such as TabLLM~\citep{hegselmann2023tabllm} have shown that pre-trained language models can leverage rich semantic priors to achieve competitive, and in some cases superior, performance on heterogeneous or context-rich datasets where conventional models struggle.

However, this paradigm shift introduces a critical challenge: \textbf{interpretability}. In high-stakes domains such as healthcare and finance, model predictions must not only be accurate but also explainable and audit-able. Stakeholders require faithful attributions that reveal which features causally influence a decision, enabling practitioners to identify biases, validate domain logic, and build trust in automated systems. While tree-based models benefit from well-established interpretability frameworks like TreeSHAP~\citep{lundberg2017unified}, LLM-based tabular classifiers remain largely opaque. Existing interpretation attempts either rely on global linear proxies such as fitting logistic regression to the model's predictions or apply token-level attribution methods designed for natural language tasks, neither of which adequately capture the instance-specific, non-linear decision logic encoded in fine-tuned LLM parameters.

The interpretability gap is further compounded by fundamental architectural differences between LLMs and traditional classifiers. Unlike discriminative models that output fixed probability vectors, generative LLMs distribute probability mass over vast vocabularies, requiring careful extraction and aggregation of class probabilities. Moreover, standard perturbation-based attribution methods operate at the sub-word token level, which is semantically inappropriate for tabular data where multi-token spans (e.g., ``Age: 50'') represent atomic feature values. Naive token deletion can corrupt prompt syntax and trigger model hallucinations unrelated to true feature importance. Finally, most existing methods measure feature impact via scalar probability shifts or prediction flips, failing to capture the model's full distributional uncertainty - a critical oversight when evaluating confidence in probabilistic decision-making.

In this work, we introduce \textbf{TabSHAP}, a model-agnostic interpretability framework specifically designed to attribute decision logic in LLM-based tabular classifiers under empirical faithfulness tests. Our approach makes three core contributions:

\textbf{1. Distributional Attribution via Jensen-Shannon Divergence.} TabSHAP measures feature importance by comparing the model's full output probability distribution before and after removing a feature, using Jensen-Shannon Divergence (JSD). Unlike methods that only track whether the predicted label changes, this captures shifts in the model's confidence across all classes, providing a more complete measure of each feature's influence.

\textbf{2. Feature-Level Atomic Masking.} In serialized tabular prompts, a single feature (e.g., ``\texttt{Age: 50}'') spans multiple tokens. Rather than masking individual tokens---which can corrupt the prompt---TabSHAP removes the entire key-value pair as one atomic unit. This preserves prompt syntax and ensures that perturbations correspond to meaningful feature removal.

\textbf{3. Verbalizer-Style Class Aggregation.} Generative LLMs may assign probability to multiple tokens that represent the same class label (e.g., ``\texttt{Yes}'', ``\texttt{yes}'', ``\texttt{ yes}''). TabSHAP aggregates mass from the top-$K$ next-token logits that match each class under a fixed label-to-token mapping before computing attributions, producing stable scores even for distilled or quantized models.

We validate TabSHAP through comprehensive experiments on the Adult Income and Heart Disease benchmarks. Deletion curve analysis demonstrates that TabSHAP achieves significantly higher faithfulness than random and tree-based baselines, with features ranked by our method causing precipitous drops in model confidence when removed. Comparative analysis against XGBoost reveals that while LLMs and tree ensembles capture similar high-level causal signals, they rely on fundamentally different decision logic---with LLMs exhibiting semantic biases toward text-rich features. Finally, we compare JSD-, KL-, and L1-based TabSHAP rankings under identical coalition sampling and deletion protocols (\S\ref{sec:experiments}).

By providing faithful, feature-level attributions for LLM-based tabular classifiers, TabSHAP enables the responsible deployment of these models in high-stakes domains, bridging the gap between the impressive generalization of language models and the interpretability requirements of critical decision-making systems.

\section{Related Work}

\paragraph{Large Language Models for Tabular Data.}
The application of Transformers to structured data has evolved from specific
architectures to the serialization of tabular data into natural language prompts
for general-purpose LLMs. \citet{hegselmann2023tabllm} formalized this with
\textbf{TabLLM}, demonstrating that fine-tuning LLMs on serialized strings
yields state-of-the-art performance, particularly on few-shot and context-heavy
tasks where traditional tree-based models struggle. Subsequent works have
further validated the superior generalization of LLMs on ``dirty'' or semantic
datasets~\citep{dinh2022lift,wu2025tablebench}. However, interpretability in this domain remains under-explored. \citet{hegselmann2023tabllm} attempted to interpret TabLLM by fitting a global Logistic Regression surrogate to the model's zero-shot predictions. While this provides a general overview of feature weights, it relies on a linear proxy that fails to capture the instance-specific, non-linear decision logic encoded in the fine-tuned LLM parameters. Our work addresses this gap by interpreting the fine-tuned model directly, rather than explaining a global proxy.

\paragraph{Feature Attribution in LLMs.}
Post-hoc attribution methods fall broadly into gradient-based and perturbation-based categories. Gradient-based methods, including saliency maps~\citep{simonyan2013deep} and Integrated Gradients~\citep{sundararajan2017axiomatic}, have been applied to LLMs but are often computationally prohibitive and noisy in deep architectures. Perturbation-based methods, specifically Shapley Values \citep{lundberg2017unified}, offer theoretically grounded axioms for attribution. \citet{horovicz2024tokenshap} and \citet{enouen2023textgenshap} recently adapted Monte Carlo Shapley estimation to LLMs, treating individual tokens as players to derive importance scores. Expanding attribution to multimodal settings, \citet{goldshmidt2025pixelshap} introduced \textsc{PixelSHAP} to reveal object-level focus in Vision-Language Models, showing the necessity of attributing semantically meaningful units (objects) rather than raw inputs (pixels). However, for tabular tasks, existing text-based methods still typically operate at the sub-word level, which is equally suboptimal since tabular semantic meaning is encapsulated in multi-token fields (e.g., ``Age: 50'' is split into four tokens). \textbf{TabSHAP} bridges this gap by enforcing feature-level token aggregation to maintain semantic integrity, drawing inspiration from higher-level semantic abstraction, and replacing standard logit-drop metrics with strict distributional measures. These advancements in interpretability are deeply necessary for building trust in LLM applications and foundational data models, aligning with emerging research priorities.

\paragraph{Interactions and Distributional Analysis.}
Recent advancements have sought to move beyond simple marginal attribution. \citet{kang2025spex} proposed a rigorous framework using sparse Fourier transforms to detect higher-order feature interactions in long-context LLMs ($n \approx 1000$). While SPEX focuses on scalability and interactions, our work targets the specific constraints of tabular classification ($n \approx 20$), where faithful \textit{marginal} attribution is the primary auditing requirement. Furthermore, while most existing methods measure impact via scalar probability shifts, our work leverages \textbf{Jensen-Shannon Divergence (JSD)}. This aligns with recent efforts in measuring diagnostic uncertainty, acknowledging that a feature's importance is defined not just by its ability to flip a label, but by its contribution to the model's distributional certainty.

\begin{figure}[H]
    \centering
    \includegraphics[width=0.5\textwidth]{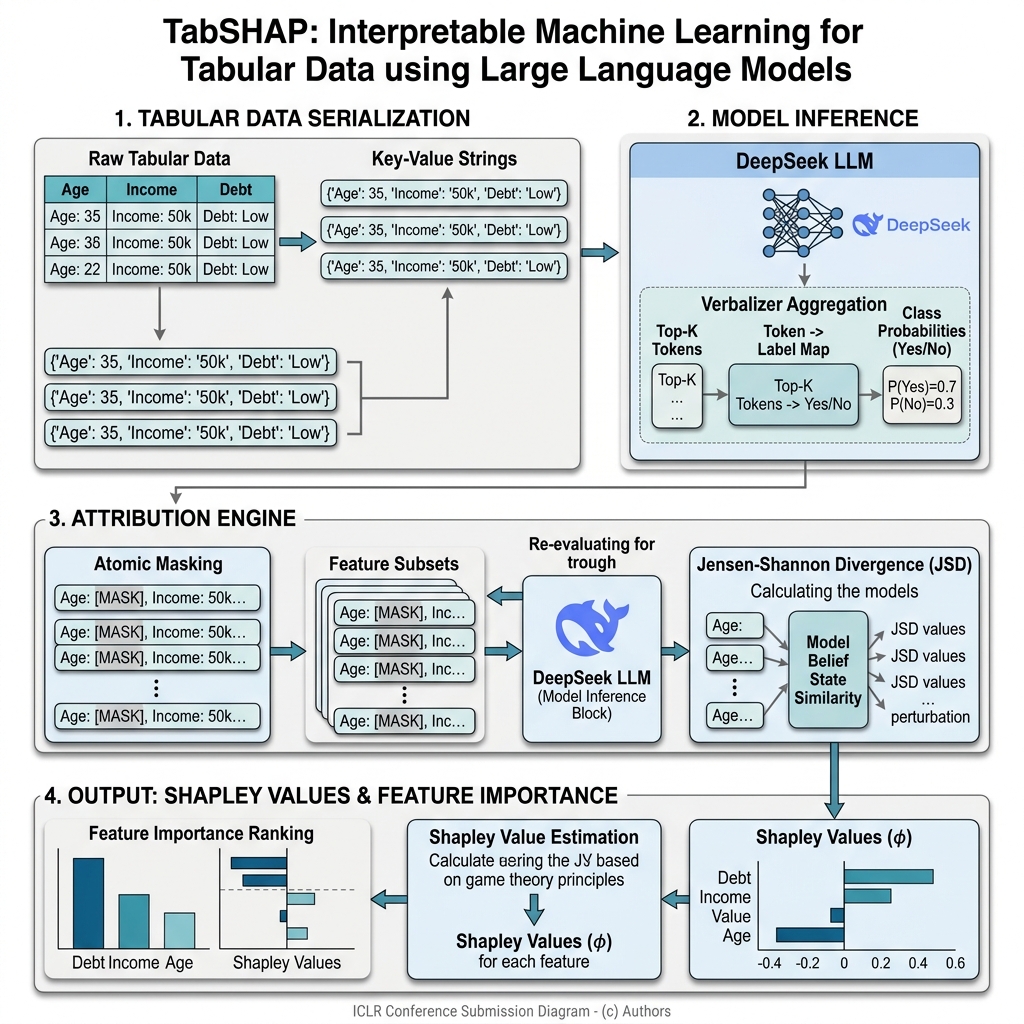}
    \caption{Workflow of \textsc{TabSHAP}: prompt serialization, LLM inference, top-$K$ class aggregation, atomic feature omission, and sampled-coalition attribution with Jensen--Shannon divergence.}
    \label{fig:tabshap_workflow}
\end{figure}

\section{Methodology}
\label{sec:methodology}
We propose \textsc{TabSHAP}, a post-hoc interpretability framework for LLMs fine-tuned on tabular classification tasks. Building on the sampled-coalition procedure introduced by TokenSHAP~\citep{horovicz2024tokenshap}, our approach adapts perturbation and scoring to enforce field-level masking and distributional comparison suited to serialized tabular prompts. Figure~\ref{fig:tabshap_workflow} summarizes the pipeline: serialization, inference, top-$K$ class aggregation, omission-based masking, Jensen--Shannon similarity, and normalized importance scores.

\subsection{Model Architecture and Optimization}
Our framework is instantiated using DeepSeek-R1-Distill-Llama-8B, a state-of-the-art distilled reasoning model chosen for its balance of inference efficiency and logic capabilities. To adapt this general-purpose model to the specific distribution of tabular risk scoring, we utilize Quantized Low-Rank Adaptation (QLoRA) \citep{dettmers2024qlora}. QLoRA allows us to fine-tune the model in 4-bit precision while freezing the base parameters, significantly reducing memory overhead while retaining high-fidelity representations. The training pipeline is accelerated using the Unsloth optimization framework, which provides optimized kernels for gradient checkpointing and faster backpropagation.

\subsection{Tabular Serialization and Fine-Tuning}
\label{sec:serialization}
Let $\mathcal{D} = \{(\mathbf{x}_i, y_i)\}_{i=1}^N$ be a tabular dataset where $\mathbf{x}_i$ contains $M$ features and $y_i \in \mathcal{C}$ is the target label (e.g., a binary or multiclass classification target). To bridge the modality gap, we employ a deterministic serialization function $\mathcal{S}(\mathbf{x})$ that converts structured rows into instruction-tuning prompts.

Unlike the natural language sentence templates often favored in prior work (e.g., TabLLM's ``The age is 39...'') \citep{hegselmann2023tabllm}, we utilize a \textbf{concise key-value representation} to maximize token efficiency and structural clarity. As implemented in our conversion script, numerical values are cast to integers, and categorical strings are normalized (lowercased, spaces replaced with underscores) to reduce tokenizer fragmentation. The features are concatenated into a space-delimited string:

\begin{equation}
    \mathcal{F}(\mathbf{x}) = \bigoplus_{j=1}^M (\texttt{key}_j\text{:}\texttt{value}_j)
\end{equation}

For example, a row is serialized as: \texttt{age:50 workclass:private ...}. This feature string is embedded into a standard instruction-tuning template.

\subsection{Output Distribution and Verbalizer Aggregation}
\label{sec:verbalizer}

Unlike discriminative classifiers that output a fixed vector of class probabilities, generative LLMs distribute probability mass over a vast vocabulary $\mathcal{V}$. To extract a calibrated classification probability $P(y|\mathbf{x})$ suitable for Shapley estimation, we perform a direct logit probe at the final token position of the prompt.

Let $\mathbf{z} \in \mathbb{R}^{|\mathcal{V}|}$ be the logits output by the model's language modeling head. We compute the probability distribution over the vocabulary via the softmax function: $P(t|\mathbf{x}) = \text{softmax}(\mathbf{z})_t$ for each token $t \in \mathcal{V}$.

\textbf{Token Aggregation:} A critical challenge in LLM interpretability is that tokenization is sensitive to spacing and casing. For example, the semantic concept ``Yes'' may be represented by distinct tokens such as ``\texttt{Yes}'', ``\texttt{ yes}'', or fragmented tokens like ``\texttt{ye}'' + ``\texttt{s}''. Relying on a single token ID causes feature attribution to be unstable. To resolve this, we define a \textbf{Verbalizer Mapping} $\mathcal{T}: \mathcal{C} \rightarrow 2^{\mathcal{V}}$ that maps each target class $c \in \mathcal{C}$ (e.g., $\{\text{Yes}, \text{No}\}$) to a set of semantically equivalent tokens.

We compute the raw probability mass for a class $c$ by aggregating over the top-$K$ most probable tokens:
\begin{equation}
    P_{\text{raw}}(c|\mathbf{x}) = \sum_{t \in \text{top-}K(P(\cdot|\mathbf{x}))} \mathbb{I}[t \in \mathcal{T}(c)] \cdot P(t|\mathbf{x})
\end{equation}
This dynamic aggregation is particularly critical for distilled or quantized models, where tokenization artifacts may cause the model to output fragmented tokens. Finally, to ensure a valid probability distribution for the Shapley calculation, we normalize the aggregated probabilities over the candidate class subspace:
\begin{equation}
    P(c|\mathbf{x}) = \frac{P_{\text{raw}}(c|\mathbf{x})}{\sum_{c' \in \mathcal{C}} P_{\text{raw}}(c'|\mathbf{x})}
\end{equation}

\subsection{Feature Perturbation via Atomic Masking}
\label{sec:perturbation}

A core challenge in LLM interpretability is defining the ``absence'' of a feature in a text stream. Naive token-level deletion (e.g., masking individual tokens like ``\texttt{5}'' in ``\texttt{Age: 50}'') corrupts the semantic integrity of the prompt, often creating invalid numerical values (e.g., ``\texttt{Age: 0}'') that trigger model hallucinations unrelated to the feature's true importance.

To address this, \textsc{TabSHAP} implements \textbf{Atomic Feature Masking}. Each tabular field is serialized as a single space-delimited token of the form \texttt{key:value} (see \S\ref{sec:serialization}); we treat each such token as one atomic unit. During coalition evaluation, if a feature is absent from coalition $S$, its entire \texttt{key:value} string is \textbf{omitted} from the input block---we do not substitute mean, mode, or placeholder values. The instruction and response template remain fixed, so the model sees a valid prompt containing only the active features in $S$.

\subsection{Distributional Similarity \& Estimation}
\label{sec:algorithm}

Computing exact Shapley values is computationally intractable for LLMs due to the exponential number of feature combinations. We adopt the Monte Carlo estimation framework from TokenSHAP \citep{horovicz2024tokenshap}, but we fundamentally alter the metric to suit tabular classification.

Instead of semantic similarity, \textsc{TabSHAP} measures the \textbf{preservation of the model's belief state}. We define the value function $v(S)$ of a feature coalition $S$ as its \textit{Distributional Similarity} to the full model output. Formally, let $P_{\text{full}}$ be the class distribution given all features, and $P_S$ the distribution under coalition $S$. Let $\mathrm{JSD}_{\mathrm{nat}}(\cdot \| \cdot)$ denote Jensen--Shannon divergence with the natural logarithm (in nats). We map divergence to a bounded similarity in $[0,1]$ by normalizing by $\ln 2$ (the maximum of $\mathrm{JSD}_{\mathrm{nat}}$ for a binary support):
\begin{equation}
    \label{eq:sim}
    v(S) = 1 - \min\left\{ \frac{\mathrm{JSD}_{\mathrm{nat}}(P_{\text{full}} \ \| \ P_S)}{\ln 2},\, 1 \right\}
\end{equation}
A value of $v(S) \approx 1$ implies that $S$ yields a class distribution close to the full-input distribution; $v(S)$ near $0$ implies large divergence. For ablations, we swap the distance in this step for KL divergence or L1 distance, using the same bounded similarity maps as in our evaluation code (\texttt{kl\_divergence}, \texttt{l1\_distance}).

Following the sampled-coalition estimator used in TokenSHAP-style methods, we \textbf{approximate} feature scores by Monte Carlo sampling over coalitions: we always include all $M$ leave-one-out coalitions, add random non-empty coalitions subject to a sampling ratio $r$ and a cap $C_{\max}$, then set $\phi_j$ to the difference between the average $v(S)$ over sampled coalitions that \textit{include} $j$ versus those that \textit{exclude} $j$. This is a tractable proxy for full Shapley values, not the exact weighted Shapley formula over all $2^M$ subsets. Finally, we shift by the minimum and normalize to sum to one. Algorithm~\ref{alg:tabshap} details the procedure.
\begin{algorithm}[t]
\caption{\textsc{TabSHAP}: Distributional Feature Attribution}
\label{alg:tabshap}
\begin{algorithmic}[1]
\Require Model $\mathcal{M}$, Input $\mathbf{x}$, Feature Set $\mathcal{F}$, \textbf{Class Set $\mathcal{K}$}, Sampling Ratio $r$, Max Coalitions $C_{\max}$
\Ensure Normalized importance scores $\phi_j$ for each feature $f_j$

\State \textbf{Initialization:}
\State Parse $\mathbf{x}$ into $M$ atomic features $\{f_1,\ldots,f_M\}$ (space-delimited \texttt{key:value} tokens)
\State $\mathbf{z}_{\text{full}} \leftarrow \mathcal{M}(\mathbf{x})$
\State $P_{\text{full}} \leftarrow \text{ClassProb}(\mathbf{z}_{\text{full}}, \mathcal{K})$ \Comment{Top-$K$ logit aggregation over $\mathcal{K}$; see \S\ref{sec:verbalizer}}

\State \textbf{Leave-one-out coalitions:}
\State $\mathcal{E} \leftarrow \{ \mathcal{F} \setminus \{f_j\} : j = 1,\ldots,M \}$ \Comment{$M$ essential coalitions}

\State \textbf{Additional random coalitions:}
\State $N_{\text{extra}} \leftarrow \big\lfloor \big( (2^M - 1) - M \big) \cdot r \big\rfloor$
\State $N_{\text{extra}} \leftarrow \min\big( N_{\text{extra}},\, \max(0,\, C_{\max} - M) \big)$
\State $\mathcal{S} \leftarrow$ Sample $N_{\text{extra}}$ distinct non-empty coalitions uniformly from the powerset (excluding members of $\mathcal{E}$)
\State $\mathcal{C} \leftarrow \mathcal{E} \cup \mathcal{S}$

\State \textbf{Evaluation loop:}
\For{each coalition $c \in \mathcal{C}$}
    \State $\mathbf{x}_c \leftarrow \text{ConstructPrompt}(c)$ \Comment{Omit absent \texttt{key:value} tokens; keep template fixed}
    \State $P_c \leftarrow \text{ClassProb}(\mathcal{M}(\mathbf{x}_c), \mathcal{K})$
    \State $\mathrm{Sim}(P_{\text{full}}, P_c) \leftarrow 1 - \min\big\{ \mathrm{JSD}_{\mathrm{nat}}(P_{\text{full}} \| P_c) / \ln 2,\, 1 \big\}$
\EndFor

\State \textbf{Attribution (sampled-coalition proxy):}
\For{$j = 1$ to $M$}
    \State $\text{with}_j \leftarrow$ Average $\mathrm{Sim}$ over $c \in \mathcal{C}$ with $f_j \in c$
    \State $\text{without}_j \leftarrow$ Average $\mathrm{Sim}$ over $c \in \mathcal{C}$ with $f_j \notin c$
    \State $\phi_j \leftarrow \text{with}_j - \text{without}_j$
\EndFor

\State \textbf{Normalization:}
\State $\phi \leftarrow \phi - \min(\phi)$
\State $\phi \leftarrow \phi / \sum_j \phi_j$
\State \textbf{return} $\phi_1, \dots, \phi_M$
\end{algorithmic}
\end{algorithm}

\section{Experiments}
\label{sec:experiments}

We evaluate \textsc{TabSHAP} on two distinct tabular tasks to demonstrate its faithfulness, alignment with established baselines, and ability to capture logical constraints in multiclass settings.

\subsection{Experimental Setup}
\textbf{Datasets.} We utilize the \textbf{Adult Income} dataset ($N=48,842$, 14 features) as a standard binary classification benchmark. We add the \textbf{Heart Disease} dataset ($N=1,025$, 13 features) to provide validation in a critical, high-stakes healthcare domain where interpretability is essential. Table \ref{tab:datasets} summarizes the datasets.

\begin{table}[h]
\centering
\caption{Summary of evaluation datasets used in experiments.}
\label{tab:datasets}
\begin{tabular}{lrrl}
\toprule
\textbf{Dataset} & \textbf{Instances} & \textbf{Features} & \textbf{Task} \\
\midrule
Adult Income & 48,842 & 14 & Income $>$\$50K \\
Heart Disease & 1,025 & 13 & Disease Prediction \\
\bottomrule
\end{tabular}
\end{table}

\textbf{Models and Hyperparameters.} We fine-tune \texttt{DeepSeek-R1-Distill-Llama-8B} \citep{touvron2023llama} using QLoRA \citep{dettmers2024qlora} on serialized feature strings. The training leverages the Unsloth optimization framework for memory efficiency. Key hyperparameters include a LoRA Rank of 16, $\alpha=16$, and 4-bit precision loading. For TabSHAP attribution, we use coalition sampling ratio $r=0.4$, a cap of $C_{\max}=800$ coalitions (including all $M$ leave-one-out sets), top-$10$ logits for class aggregation, and deterministic probing at the final prompt position (no stochastic decoding), so differences across runs come from feature subsets rather than sampling noise.

\textbf{Baselines.} We compare \textsc{TabSHAP} against:
\begin{itemize}
    \item \textbf{Random Removal:} A baseline where features are masked in random order.
    \item \textbf{XGBoost + TreeSHAP:} A state-of-the-art tree ensemble interpreted via TreeSHAP \citep{lundberg2017unified}, serving as a "proxy ground truth" for tabular logic.
\end{itemize}

\subsection{Faithfulness Evaluation (Deletion Curve)}
\label{sec:faithfulness_eval}
We assess whether \textsc{TabSHAP} identifies features that genuinely steer the model by \textbf{sequentially deleting} atomic \texttt{key:value} tokens from the \texttt{\#\#\# Input:} block (instruction and response template unchanged). At each step we recompute the class distribution and track the \textbf{probability mass on the originally predicted class} (top-$K$ aggregation). We report the mean over test instances versus the fraction of features removed (normalized by the average number of features per instance). Removal orders compared are: (i) JSD-based TabSHAP ranking, (ii) XGBoost precomputed TreeSHAP ranking (when available), and (iii) a random permutation. This yields the faithfulness plot (JSD TabSHAP vs.\ baselines) in our evaluation pipeline.

\begin{figure}[h]
    \centering
    \begin{minipage}{0.48\textwidth}
        \centering
        \includegraphics[width=\linewidth]{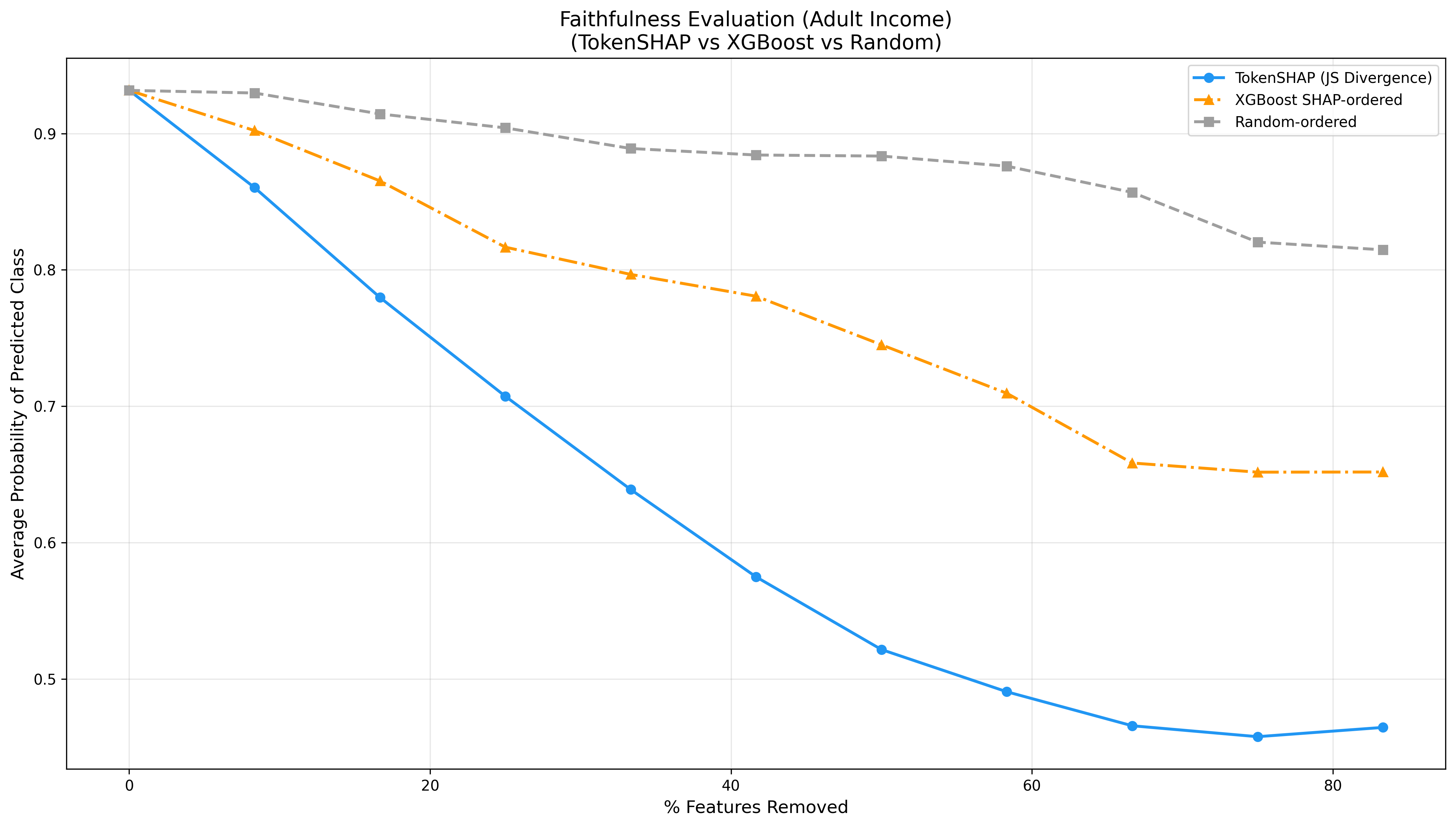}
        \caption{\textbf{Faithfulness on Adult Income.} \textsc{TabSHAP} achieves a sharp confidence drop under ordered feature deletion.}
        \label{fig:faithfulness_income}
    \end{minipage}\hfill
    \begin{minipage}{0.48\textwidth}
        \centering
        \includegraphics[width=\linewidth]{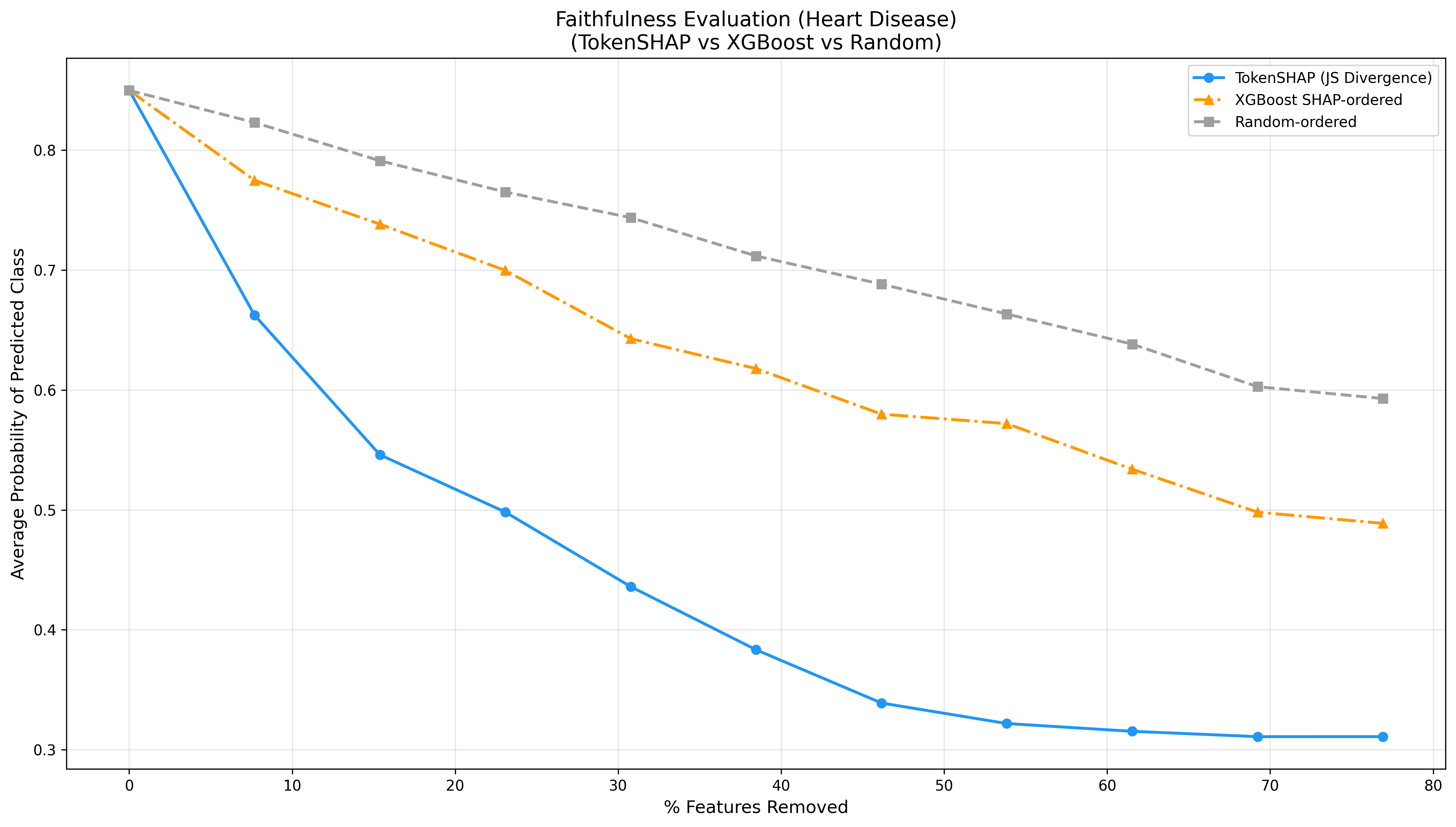} 
        \caption{\textbf{Faithfulness on Heart Disease.} \textsc{TabSHAP} isolates diagnostic features under the same deletion protocol.}
        \label{fig:faithfulness_heart}
    \end{minipage}
    \label{fig:faithfulness_experiments}
\end{figure}

\subsection{Distance Metric Ablation Study}
\label{sec:metric_ablation}
\begin{figure}[h]
    \centering
    \begin{minipage}{0.48\textwidth}
        \centering
        \includegraphics[width=\linewidth]{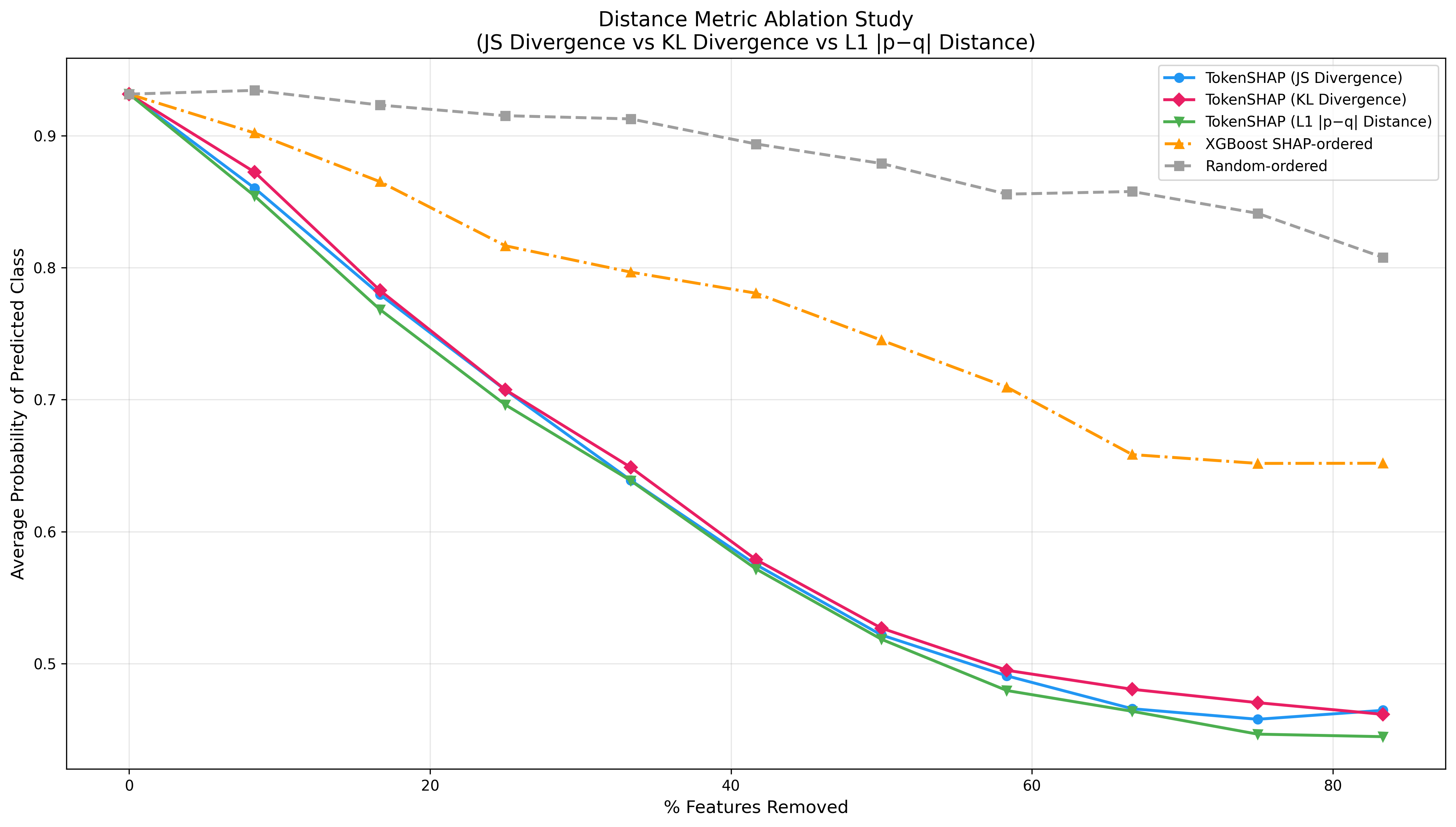} 
        \caption{\textbf{Adult Income: distance-metric ablation.} Deletion curves for JSD-, KL-, and L1-based TabSHAP vs.\ baselines.}
        \label{fig:ablation_income}
    \end{minipage}\hfill
    \begin{minipage}{0.48\textwidth}
        \centering
        \includegraphics[width=\linewidth]{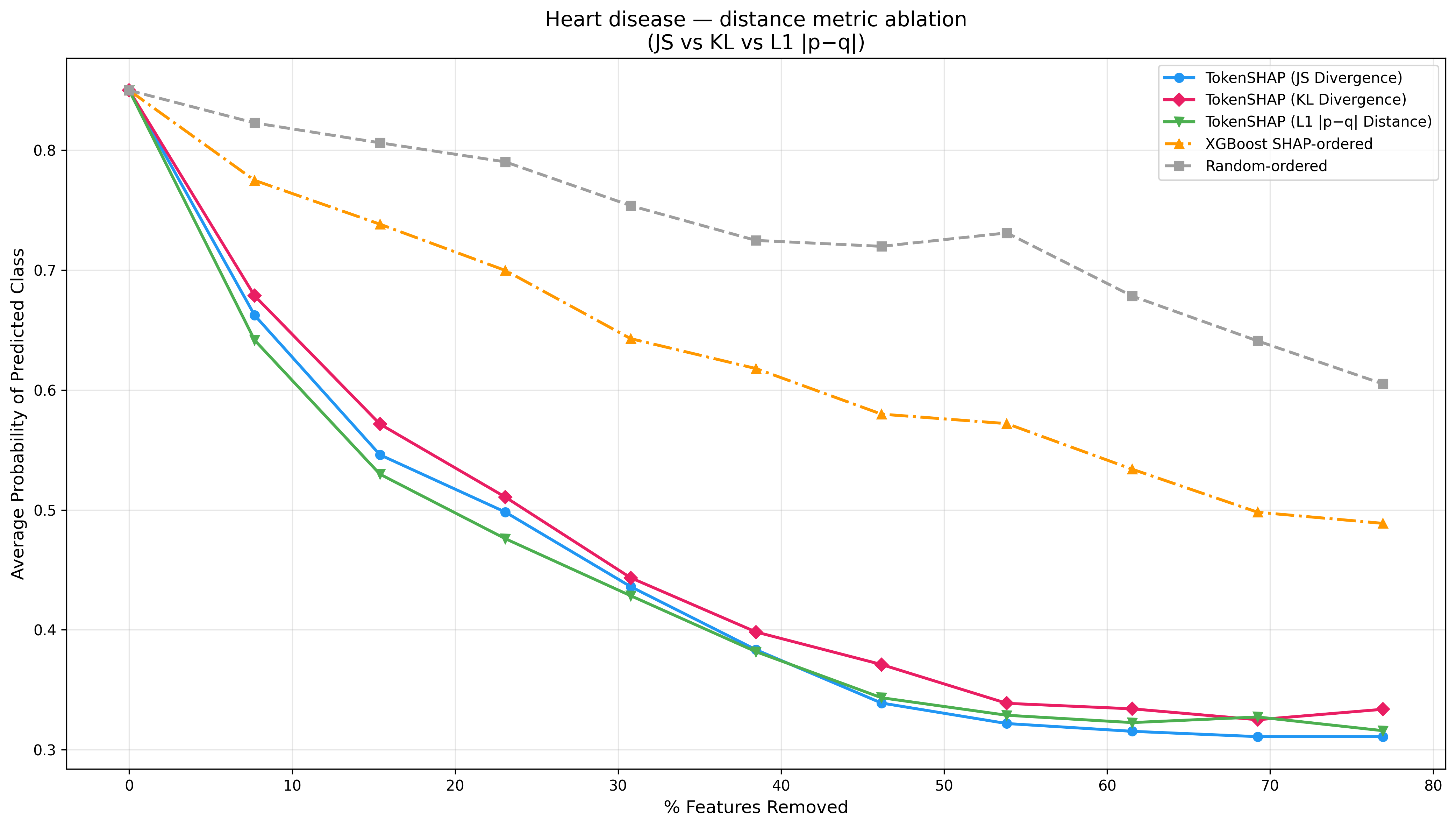} 
        \caption{\textbf{Heart Disease: distance-metric ablation.} Same comparison on the Heart benchmark.}
        \label{fig:ablation_heart}
    \end{minipage}
    \label{fig:ablation_experiments}
\end{figure}

We hold the test subset and TabSHAP hyperparameters fixed, load JSD attributions from cache, and compute KL- and L1-based TabSHAP rankings on the \textbf{same} indices (caching each metric's rankings after the first run). We then run the \emph{identical} deletion procedure as in \S\ref{sec:faithfulness_eval}, but compare removal curves induced by JSD-, KL-, and L1-based orderings, alongside XGBoost- and random-order baselines. Up to $10$ features are removed per instance (or fewer if fewer fields exist). This produces the metric-ablation curves used in Figure~\ref{fig:ablation_experiments}.

\subsection{Comparative Analysis: LLM vs. XGBoost}
We compare the global feature rankings derived from \textsc{TabSHAP} against those from XGBoost (TreeSHAP) on the Adult Income dataset. 

\textbf{Rank Correlation.} We observe a moderate Spearman rank correlation ($\rho \approx 0.65$) between the two methods. This suggests that while the fine-tuned LLM captures the fundamental causal signals (e.g., \texttt{Capital Gain}, \texttt{Marital Status}), it relies on different internal logic than the tree ensemble. 

\textbf{Semantic Bias.} Qualitative analysis reveals that the LLM tends to assign disproportionately higher importance to semantic, text-rich features (e.g., \texttt{Occupation}, \texttt{Education}) compared to XGBoost, which prioritizes precise numerical splits (e.g., \texttt{Hours-per-week}). This distinction highlights the value of using a faithful interpreter like \textsc{TabSHAP} rather than a global proxy; it reveals that Tabular LLMs solve tasks by leveraging semantic priors, a behavior concealed when approximating the LLM with a linear model.

\section{Limitations}
While TabSHAP provides faithful, instance-level attributions, several limitations should be noted. First, coalition sampling and the with/without score (Algorithm~\ref{alg:tabshap}) approximate Shapley-style importance rather than the full exponential-time Shapley formula; faithfulness depends on $r$, $C_{\max}$, and the omission (not imputation) semantics. Second, repeated forward passes per instance make the method expensive in wall-clock time, though sampling caps cost in practice. Third, our evaluation is limited to two benchmarks with small $M$; high-dimensional tabular settings remain open. Fourth, we report marginal attributions and do not model explicit feature interactions. Fifth, class probability extraction uses top-$K$ logit matching to predefined labels, which may need tuning for new tasks or vocabularies. Finally, we validate on a single fine-tuned backbone (DeepSeek-R1-Distill-Llama-8B); other architectures may behave differently.

\section{Conclusion}
In this work, we introduced \textsc{TabSHAP}, a model-agnostic interpretability framework for LLMs on serialized tabular prompts. Using Jensen--Shannon divergence (with KL/L1 ablations in the same implementation), atomic omission of \texttt{key:value} fields, and top-$K$ logit class aggregation, we measure local feature importance and validate it with deletion curves (predicted-class probability vs.\ fraction of features removed). Faithfulness evaluation compares JSD TabSHAP to XGBoost- and random-order removal; the metric study compares JSD, KL, and L1 TabSHAP orderings on matched test instances with cached attributions. Future work includes larger $M$, interaction effects, and additional backbones.

\bibliography{iclr2026_conference}
\bibliographystyle{iclr2026_conference}

\appendix

\section{Prompt Serialization Formatting}
\label{app:serialization}
To ensure robust tabular ingestion by the LLM, we employed a deterministic serialization script. Numerical integers were passed directly as text, while categorical values were normalized by lowercasing and replacing whitespace with underscores. This mitigated tokenizer fragmentation. Full schema translations were embedded into the standard DeepSeek instruction-tuning template, encapsulated between strict \texttt{\#\#\# Input:} and \texttt{\#\#\# Response:} markers to guarantee the attribution engine selectively masked correct feature bounds.

\section{Hyperparameters and Optimization}
\label{app:hyperparameters}
All \textsc{TabSHAP} models were built upon \texttt{DeepSeek-R1-Distill-Llama-8B}. To manage memory footprint while capturing complex tabular logic, models were fine-tuned via QLoRA.
\begin{itemize}
    \item \textbf{LoRA Configuration:} Rank $r = 16$, Alpha $\alpha = 16$, Target Modules included query, key, value, and output projections.
    \item \textbf{Optimization:} Unsloth kernel acceleration applied for 4-bit loading quantization.
    \item \textbf{TabSHAP / ablation:} $r=0.4$, $C_{\max}=800$, top-$10$ class logits, up to $10$ sequential removals per curve; JSD rankings loaded from \texttt{tokenshap\_validation\_cache.json}; KL/L1 rankings written to separate JSON caches on first run so all metrics use the same \texttt{selected\_test\_indices}.
\end{itemize}

\section{Dataset Dimensions}
\label{app:datasets}
\textbf{Adult Income.} Comprises $N = 48,842$ instances. Key text features include \texttt{workclass}, \texttt{education}, \texttt{marital\_status}, \texttt{occupation}, and \texttt{race}. Key numerical parameters include \texttt{age}, \texttt{capital\_gain}, and \texttt{hours\_per\_week}. Target threshold is income $>$\$50K.

\textbf{Heart Disease.} Comprises 13 critical clinical parameters encompassing categorical inputs (e.g., chest pain type, resting electrocardiographic results) and continuous readings (e.g., resting blood pressure, serum cholesterol). Target represents heart disease diagnosis.

\end{document}